\documentclass[lettersize,journal]{IEEEtran}
\usepackage{amsmath,amsfonts}
\usepackage{algorithmic}
\usepackage{array}
\usepackage[caption=false,font=normalsize,labelfont=sf,textfont=sf]{subfig}
\usepackage{textcomp}
\usepackage{stfloats}
\usepackage{url}
\usepackage{verbatim}
\usepackage{graphicx}
\usepackage{xcolor}
\def\BibTeX{{\rm B\kern-.05em{\sc i\kern-.025em b}\kern-.08em
    T\kern-.1667em\lower.7ex\hbox{E}\kern-.125emX}}
\usepackage{balance}
\usepackage{subcaption}
\usepackage{pifont}
\usepackage{multirow}
\usepackage{makecell} 

\begin{document}

\title{Object-level Cross-view Geo-localization with Location Enhancement and Multi-Head Cross Attention}

\author{Zheyang Huang, Jagannath Aryal, Saeid Nahavandi, Xuequan Lu, Chee Peng Lim, Lei Wei, Hailing Zhou

\thanks{Zheyang Huang is with the Meitu Inc., China (email: jasonhuang1999cn@gmail.com).

Jagannath Aryal is with University of Melbourne, VIC, Australia (e-mail: jagannath.aryal@unimelb.edu.au), VIC, Australia.

Saeid Nahavandi and Chee Peng Lim are with Swinburne University of Technology, VIC, Australia. (emails: snahavandi@swin.edu.au and cplim@swin.edu.au)

Xuequan Lu is with The University of Western Australia, VIC, Australia. (email: bruce.lu@uwa.edu.au)

Lei Wei is with IISRI, Deakin University, VIC, Australia. (email: lei.wei@deakin.edu.au)

Hailing Zhou is the corresponding author. She is with Swinburne University of Technology, VIC, Australia. (email: hailingzhou@swin.edu.au)}}


\maketitle

\begin{abstract}

Cross-view geo-localization determines the location of a query image, captured by a drone or ground-based camera, by matching it to a geo-referenced satellite image. While traditional approaches focus on image-level localization, many applications, such as search-and-rescue, infrastructure inspection, and precision delivery, demand object-level accuracy. This enables users to prompt a specific object with a single click on a drone image to retrieve precise geo-tagged information of the object. However, variations in viewpoints, timing, and imaging conditions pose significant challenges, especially when identifying visually similar objects in extensive satellite imagery. To address these challenges, we propose an Object-level Cross-view Geo-localization Network (OCGNet).  It integrates user-specified click locations using Gaussian Kernel Transfer (GKT) to preserve location information throughout the network. This cue is dually embedded into the feature encoder and feature matching blocks, ensuring robust object-specific localization. Additionally, OCGNet incorporates a Location Enhancement (LE) module and a Multi-Head Cross Attention (MHCA) module to adaptively emphasize object-specific features or expand focus to relevant contextual regions when necessary. OCGNet achieves state-of-the-art performance on a public dataset, CVOGL. It also demonstrates few-shot learning capabilities, effectively generalizing from limited examples, making it suitable for diverse applications (https://github.com/ZheyangH/OCGNet).

\end{abstract}


\section{Introduction}
\IEEEPARstart{C}{ross-view} geo-localization allows a system to determine the geographic location of a query image-whether captured by a drone or ground-based camera—by matching it to geo-tagged reference data, such as a satellite image. It recently receives increasing attention across diverse fields, including autonomous driving \cite{ref3,ref12}, drone navigation \cite{ref4,ref13,ref14}, augmented reality \cite{ref2,ref17}, and social media \cite{ref1}. While GPS devices offer location estimates with position errors ranging from 2 to 15 meters \cite{lin2022joint}, cross-view geo-localization has potentials to provide a more accurate alternative by leveraging detailed visual information for precise localization.

\begin{figure}[t]
  \centering
  \includegraphics[width=1\linewidth]{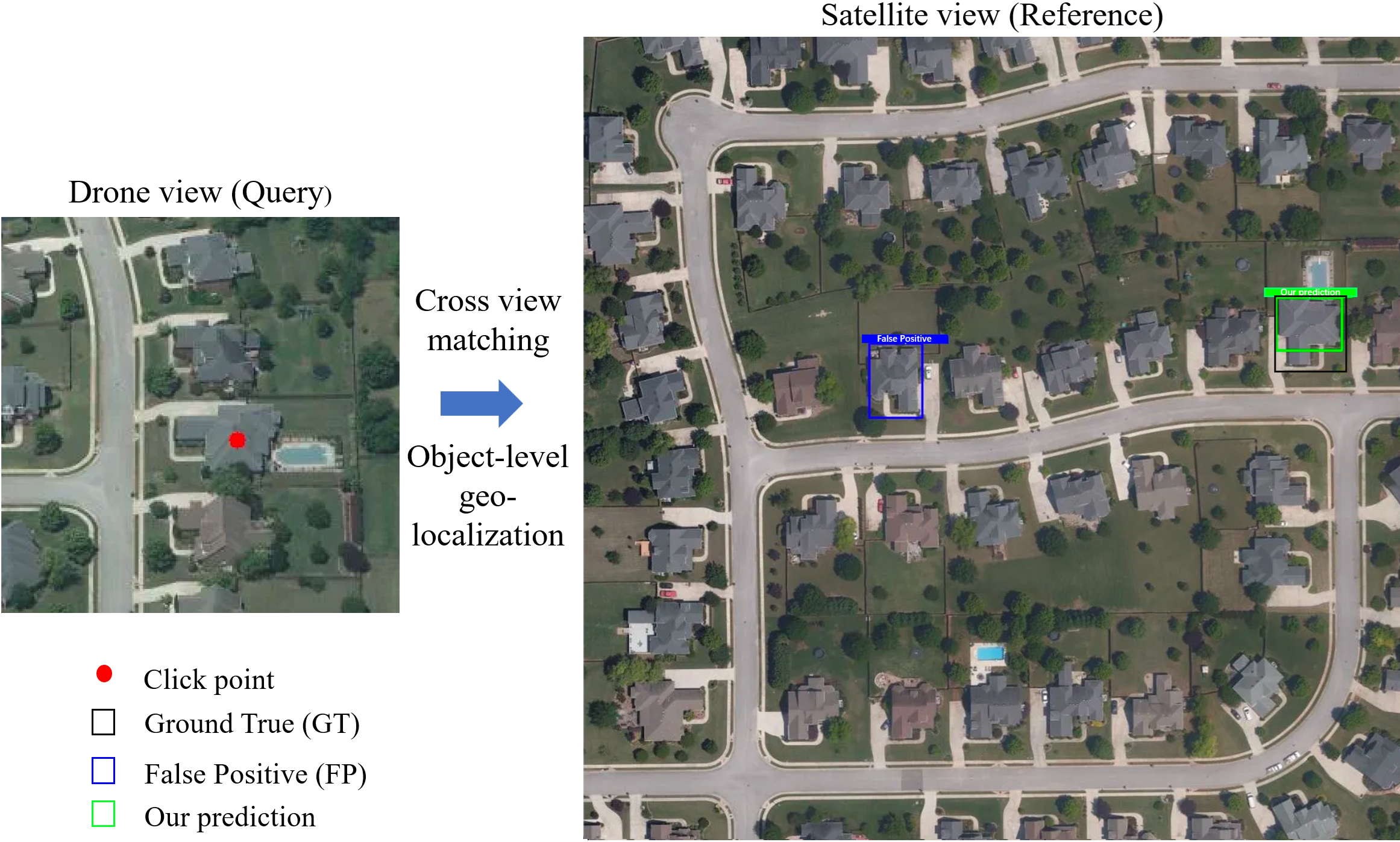}
  \caption{ Illustration of object-level geo-localization across drone and satellite views. The task is particularly challenged by targets sharing similar appearances such as houses marked by blue and black bounding boxes. }
  \label{f1}
\end{figure}

Most efforts have focused on \textbf{image-level} geo-localization. These methods treat cross-view geo-localization as an image retrieval task, identifying a geographic location of the whole reference view \cite{ref6, zheng2020university, zhu2021vigor, lin2022joint, ref7}. As the demand for fine-grained, highly accurate localization increases \cite{ref8, ref7, sun2023cross}, image-level localization is insufficient, especially for tasks of search-and-rescue missions, infrastructure inspection, event detection, and accurate delivery services. In this work, we focus on\textbf{ object-level} geo-localization that allows users to specify a target object in a query image (captured by a UAV or ground camera) and localize it within a satellite image \cite{zhu2021vigor}. Compared with image-level tasks, higher localization precision is required, where satellite imagery often covers vast areas filled with numerous objects, making it difficult to isolate and identify specific targets. This is further challenged by dealing with visually similar objects, such as houses, buildings, roundabouts or roads, as shown in Figure~\ref{f1}. Additionally, existing datasets like CVUSA \cite{ref9}, CVACT \cite{ref10}, GRAL \cite{zhu2023sues}, and University-1652 \cite{zheng2020university} are primarily for image retrieval tasks. The only available dataset for object-level geo-localization is CVOGL \cite{sun2023cross}, where query images are marked with click points to specify an object and reference images are annotated with bounding boxes to provide groundtruth detection.

{We propose OCGNet, a novel end-to-end architecture for object-level geo-localization. Unlike existing methods that integrate click-point inputs early leading to a loss of object-specific details, OCGNet introduces a Location Enhancement (LE) module that incorporates user inputs at both early and late stages. Early-stage positional embeddings provide spatial priors, while the LE module reinforces location cues post-semantic alignment, preserving spatial fidelity throughout the hierarchical fusion process \cite{kirillov2023segment, ravi2024sam}.}

{To better leverage user-provided click-points, we propose a novel Gaussian Kernel Transfer (GKT) embedding to replace the traditional Euclidean distance map \cite{sun2023cross}. GKT models click locations using a differentiable Gaussian kernel, producing spatially focused and smoothly decaying attention maps. Unlike Euclidean maps that can activate distant areas and reduce precision, GKT concentrates gradients around the target, enhancing spatial accuracy and robustness—particularly for small or ambiguous objects under large viewpoint shifts (e.g., Drone $\rightarrow$ Satellite).}

To further improve query feature quality, we introduce a learnable Multi-Head Cross Attention (MHCA) module that jointly processes query and reference images. MHCA adaptively refines query features by emphasizing distinct objects or relevant context, allowing selective attention to key regions and suppressing distractors. This promotes better object-context alignment and improves localization accuracy in complex scenarios.

OCGNet sets a new benchmark in object-level geo-localization with strong few-shot performance. Our main contributions are:
\begin{itemize}
    \item A dual-stage integration scheme that embeds click-point information early and late, preserving spatial cues throughout.
    \item A GKT-based embedding that enhances spatial focus and fine-grained feature retention.
    \item A context-aware MHCA module for adaptive query refinement against reference imagery.
    \item State-of-the-art results on standard and few-shot benchmarks, demonstrating robustness and generalization.
\end{itemize}

\section{Related work}
\subsection{Cross-view Geo-localization}
The introduction of cross-view datasets such as CVUSA \cite{ref9, zhai2017predicting}, CVACT \cite{liu2019lending}, and University-1652 \cite{zheng2020university} has significantly advanced deep learning-based vehicle geo-localization in GPS-denied environments. These methods typically formulate the localization problem as an image retrieval task, matching a ground-view image to satellite patches. Despite their success, bridging the domain gap between ground and satellite views remains a key challenge. To address it, Siamese networks have been widely used for learning cross-view similarities. CVM-Net \cite{hu2018cvm} introduced a Siamese alignment framework with location-based descriptors, while SAFA \cite{shi2019spatial} enhanced performance using polar transformations and spatial-aware embeddings, achieving strong results on CVUSA and CVACT.

Beyond those ground-satellite tasks, drone-satellite geo-localization has received considerable attention \cite{ref13, ref14, ref15, ref16, zhu2023sues}. Zheng et al. propose the University-1652 dataset \cite{zheng2020university} with drone, ground, and satellite views, establishing a benchmark using instance loss for cross-view alignment. SUES-200 \cite{zhu2023sues} extends this with multi-height drone images, diverse scenes, and realistic lighting, making it more representative than its predecessor. Previous research primarily focused on vehicle geo-localization until the CVOGL dataset \cite{sun2023cross} shifts the focus. The CVOGL dataset includes drone-satellite and ground-satellite cross-view images with click-point prompts, enabling the detection of objects in satellite images through a click-point on the query image for geo-localization.

Most recently, research has increasingly emphasized fine-grained geo-localization \cite{ref8}, which is critical for applications like autonomous navigation. Lin et al. \cite{lin2022joint} proposed a keypoint-guided coarse-to-fine matching strategy, while others introduced a square-ring partition approach to leverage spatial context \cite{ref7}. Sun et al. \cite{sun2023cross} presented an innovative object detection framework for cross-view geo-localization that encodes click-point information (identifying a target object) within the query image, fusing it with the reference image to locate the object’s bounding box. While effective, the early-stage position embedding and non-learnable fusion mechanisms continue to present challenges described earlier. This work proposes new techniques to overcome these limitations, advancing the capabilities of fine-grained cross-view geo-localization.

\subsection{ Click-point Embedding }

In the area of click-point embedding, recent developments such as SAM \cite{kirillov2023segment} and SAM2 \cite{ravi2024sam} have introduced a prompt encoder paradigm that utilizes convolution and concatenation to effectively extract features from the click-point prompt. Notably, embedding prompt information in a later stage, such as the decoder layer, has shown promising results in capturing fine-grained, localized details.  \cite{sun2023cross} employs a Euclidean distance matrix \cite{dokmanic2015euclidean} alongside concatenation to encode an object’s positional information and embed it within the query image. This approach leverages spatial relationships to reinforce the model's ability to localize targets.

Building on prior methods, we introduce GKT to provide more precise and detailed location encoding, which is further combined with a late-stage embedding strategy to preserve spatial features throughout the downstream matching process.

\subsection{Feature Matching}

Feature matching in cross-view geo-localization usually is achieved by using Siamese-based networks to measure similarity between views and then localize regions with the highest match scores \cite{liu2019lending, shi2019spatial, ref11}. To bridge the view gap, recent methods incorporate attention mechanisms that fuse satellite features with attention weights to emphasize likely target regions. For example, \cite{shi2019spatial} introduced spatial-aware feature aggregation, while \cite{sun2023cross} applied spatial attention to enhance focus on probable object locations.

Traditional attention mechanism uses efficient operations like dot products and element-wise multiplication. However, as shown in cross-modal tasks (e.g., vision-language), these methods can suffer from information loss and weak feature alignment. Advances like MHCA \cite{vaswani2017attention, chen2021crossvit} address these issues by projecting features into a shared space for selective matching. Recent works \cite{shen2024icafusion, xu2023protst, liu2024towards} validate MHCA’s effectiveness in aligning diverse modalities and maintaining robust associations under viewpoint and appearance variations.

Building on these insights, we adopt MHCA to connect query and reference views, directing attention toward the target object and relevant context. Its multi-head structure captures diverse object regions simultaneously, enabling stronger feature fusion. Attention map visualizations on the CVOGL dataset confirm the improved attention and matching precision over traditional methods.

\begin{figure*}[ht]
\centering
\includegraphics[scale=0.38]{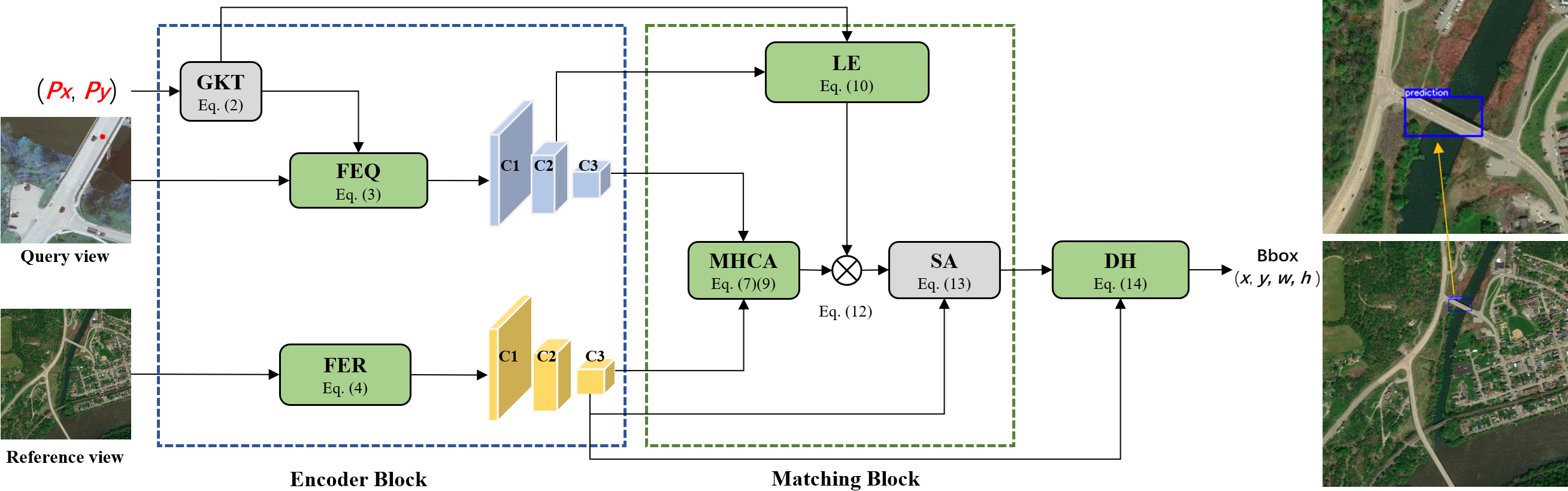}
\caption{Overview of our proposed framework where the non-learnable and learnable processes are represented by gray and green rectangles, respectively. The framework is composed of the feature Encoder Block, the feature Matching Block, and the Detection Head (DH) where the encoder block includes the Gaussian Kernel Transformation (GKT) module, Feature Extraction module for the query image (FEQ), and Feature Extraction module for the reference satellite image (FER), and the matching block incorporates the Location Enhancement (LE) module, Multi-Head Cross Attention (MHCA) and Spatial Attention (SA) modules.}
\label{f2}
\end{figure*}

\section{The proposed methodology}

The object-level cross-view geo-localization task is defined as follows: \textit{given a click-point prompt on an object in a drone or ground-level image, the goal is to detect and localize the object with a bounding box in the corresponding satellite image}. The overview of our proposed framework is shown in Fig. \ref{f2}.

\subsection{Feature Encoder}

For object-level geo-localization, the inputs consist of a query image $U$ with a given click point $P$ to specify the object of interest, along with a reference satellite image $S$. The feature encoder block is responsible for encoding both the query and satellite images, as well as integrating the click point information.

In this work, we apply a Gaussian Kernel Transformation (GKT) to encode the click point. GKT models localized attention using an exponential decay function, ensuring that nearby regions receive significantly higher focus while attention to distant areas is naturally diminished. Unlike traditional Euclidean distance maps, GKT adaptively controls attention distribution across different images, maintaining strong focus on relevant local regions while suppressing irrelevant distant ones.
{
\begin{equation}
\begin{split}
\text{M}(i, j) &= \exp\left(-\frac{(i - P_{x})^2 + (j - P_{y})^2}{2\sigma_n^2}\right), \\[6pt]
\end{split}
\end{equation}
\begin{equation}
\sigma_n = \sigma \times \sqrt{H_U^2 + W_U^2},
\end{equation}}
where \textit{P}\textsubscript{\textit{x}} and \textit{P}\textsubscript{\textit{y}} represent the $x$ and $y$ coordinates of the click point. $(i, j)$ are the coordinates of an image pixel. \text{M}$(i, j)$ is the embedding map of the click point, calculated by applying a Gaussian kernel on the domain of $U$. $\sigma_n$ is the normalized standard deviation. In Eq. (2), $H_U$ and $W_U$ are the height and width of the query image $U$. $\sigma$ is the standard deviation of the Gaussian distribution. Different settings of $\sigma$ have been tested in our experiments. {$\sigma=0.075$} and $\sigma=0.15$ work well for a drone-based query and a ground-based query, respectively.

As shown in Fig.~\ref{f2}, FEQ and FER represent the image encoders of $U$ and $S$, defined as follows.

\begin{equation}
\begin{aligned}
\text{F}_u^{C2},\text{F}_u^{C3} &= \theta(\textit{CBR}(U \oplus \text{M})),
\end{aligned}
\end{equation}
\begin{equation}
\begin{aligned}
\text{F}_s^{C3} &= \textit{CBR}(\omega(S)),
\end{aligned}
\end{equation}
where ${C3}$ denotes the final layer of an image encoder, with outputs F$_u^{C3}$ and F$_s^{C3}$ capturing high-level features for $U$ and $S$, respectively. The ${C2}$ layer represents the half-way layer of an image encoder where output features (i.e. F$_u^{C2}$) retain more spatial detail.  $\textit{CBR}$ stands for a Convolution layer followed by Batch normalization and a ReLU function as in \cite{sun2023cross}, primarily to enhance training convergence by stabilizing feature distributions. Specifically, in (3) and (4), $\textit{CBR}$ is utilized to align features from different networks into a more compatible feature space, optimizing them for subsequent MHCA fusion. {$\theta$ and $\omega$ denote feature extraction backbones based on ResNet18 \cite{he2016deep} and DarkNet \cite{redmon2018yolov3}, respectively. For ResNet18, we retain only the convolutional backbone for hierarchical feature extraction, removing the global average pooling and fully connected layers. For YOLOv3, we use the full original configuration, including DarkNet-53’s residual blocks and multi-scale detection heads.} The symbol $\oplus$ represents concatenation. {To effectively integrate the click-point information (i.e., $P$) into the three-channel query image $U$, the single-channel map generated by GKT (i.e., M$(i,j)$ in Eq.(2)) is concatenated with $U$ at an early stage, resulting in a four-channel feature representation, followed by the CBR operation.}

\begin{figure*}[ht]
\centering
\includegraphics[scale=0.40]{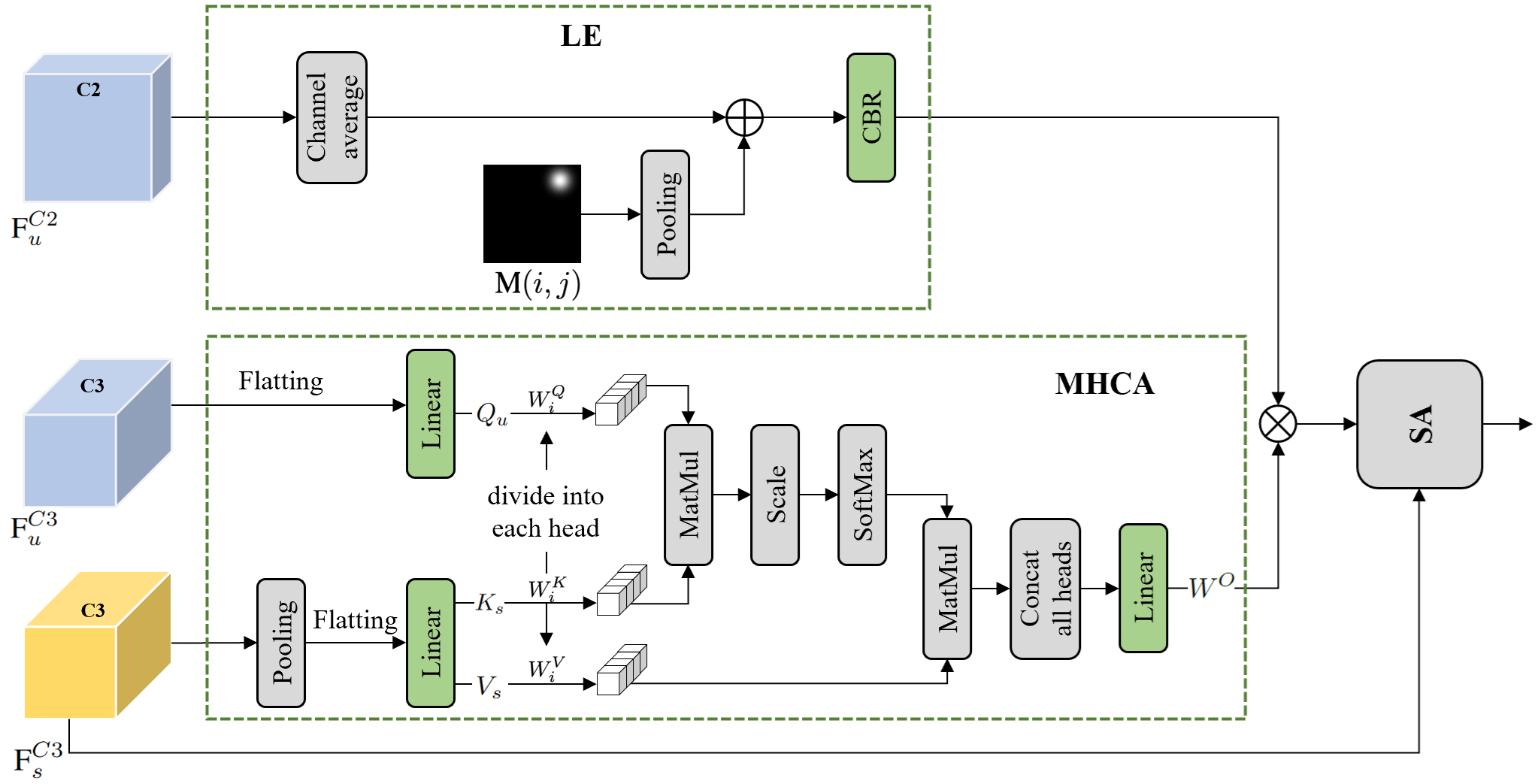}
\caption{The details of the feature matching block. The blue cubes, C2 and C3, represent outputs of the feature encoder of query \textbf{FEQ}, denoted as F$_u^{C2}$ and F$_u^{C3}$. The yellow cube, C3, represents outputs of the feature encoder of reference \textbf{FER}, denoted as F$_s^{C3}$. {The linear layer $W^O$ is used to re-integrate the outputs from all attention heads.}}
\label{f3}
\end{figure*}

\subsection{Feature Matching}
{Following the feature extraction process, the next step is to establish correspondences between the high-level representations of the query and reference images.} In feature matching, most existing methods rely on dot products and element-wise multiplication to calculate similarity between query and reference high-level features, typically without using learnable parameters or incorporating location information. This approach often struggles in challenging scenarios, such as when the reference image contains many similar objects. Additionally, object-level geo-localization for UAVs requires a lightweight matching block due to limited computational resources. To address this, we developed a three-channel input cross-view matching block with details shown in Fig. \ref{f3}. To boost performance, we introduce a multi-head cross attention module and a location enhancement module (i.e. MHCA and LE, respectively in Fig.~\ref{f2}) to enhance query features. These enhancements enable high matching accuracy using minimal learnable parameters, allowing real-time applications while effectively preserving object-specific and contextual information.

\subsubsection{MHCA: selective focuses within the query domain}
The MHCA module is to find a common space where similarities and dissimilarities between two features (i.e. F$_u^{C3}$ and F$_s^{C3}$) can be well reflected. To achieve it , we firstly transfer the encoded features to the corresponding feature vectors (i.e. Query Q, Key K, and Value V), shown as $Q_u$, $K_s$, and $V_s$. 

\begin{equation}
\begin{aligned}
Q_u &= Linear(Flat(\text{F}_u^{C3})),\\
K_s &= Linear(Flat(AvgPooling(\text{F}_s^{C3}))),\\
V_s &= Linear(Flat(AvgPooling(\text{F}_s^{C3}))),
\end{aligned}
\end{equation}
where $Flat$ is the flatting function and $Linear$ is a learnable linear projector. The query, key and value represent as $Q_u \in \text{R}^{N_c\times (H_{u} W_{u})}$, $K_s \in \text{R}^{N_c\times (H_{s}W_{s})}$ and $V_s \in \text{R}^{N_c\times (H_{s}W_{s})}$ respectively, with a channel dimension $N_c$ and an image dimension, $H$ and $W$. {Noticed that $AvgPooling$ is applied on F$_s^{C3}$ primarily to reduce computational requirements and resolution-aware scale ensures spatial compatibility between feature maps before attention computation.}

The common space of MHCA is defined based on attentions, usually calculated by: 
\begin{equation}
\begin{aligned}
    \text{Attention}(Q, K, V) &= \text{softmax} (\frac{QK^T}{\sqrt{d_k}})V
\end{aligned}
\end{equation}
where $K^T$ is the transpose of the key matrix. In our task, the attention is further formulated as: 
\begin{equation}
\begin{aligned}
    \text{F}^\text{MHCA} &= \text{Concat}(\text{head}_1, \dots, \text{head}_n) W^O,
\end{aligned}
\end{equation}
\begin{equation}
\begin{aligned}
    \text{head}_i &= \text{Attention}(Q_u W_i^Q, K_s W_i^K, V_s W_i^V).
\end{aligned}
\end{equation} 
{$W^O \in \text{R}^{nd_v\times N_c}$ is an independent linear projection matrix that combines the outputs from all attention heads and then scale it back to an original dimension. $W_i^Q, W_i^K \in \text{R}^{N_c\times d_{\text{k}}}$ and $W_i^V \in \text{R}^{N_c\times d_{\text{v}}}$ are the three projections used for the $i$th head. $d_k$ denotes the dimension of the projected query and key vectors used in the attention computation, $d_v$ is the dimension of the V vector, and $n$ is the total number of attention heads. In our experiments, the settings of $h$ = 8 and $d_k$ = $d_v$ = 64 work well. }

{To adapt to the CVOGL task, the MHCA-aligned query-reference attention map is employed to enhance high-level query features. Once the desired cross-view attentions are obtained, the next step is to emphasize these attentions within the query domain through an element-wise product operation:}
\begin{equation}
\begin{aligned}
    \text{F}_u^{E} &= \text{F}^\text{MHCA} \cdot \text{F}_u^{C3}.
\end{aligned}
\end{equation}
F$_u^{E}$ represents an enhanced query feature that properly weighted by similarities between the query and the reference.  

\subsubsection{LE: a late-stage location embedding}

The LE module is to enhance the object-specific information during feature matching, avoiding the loss of the click-point information. A late-stage embedding strategy is applied through concatenating the click-point information (i.e. M$(i,j)$ in Eq.(2)) with the low-level features of $U$ (i.e.  F$_q^{C2}$ in Eq. (3)), as follows.
\begin{equation}
\begin{aligned}
\text{F}_k^{L} &= \textit{CBR}(AvgPooling(\text{M}) \oplus \hat{\text{F}}_u^{C2})),
\end{aligned}
\end{equation}
\begin{equation}
\begin{aligned}
\hat{\text{F}_u^{C2}} &= ChannelGlobalAverage(\text{F}_u^{C2}),
\end{aligned}
\end{equation}
where we use F$_q^{C2}$ instead of F$_q^{C3}$ because F$_q^{C2}$ can capture more fine-grained spatial information than F$_q^{C3}$. Noticed that $AvgPooling$ and $ChannelGlobalAverage$ are applied on M and F$_u^{C2}$ respectively before concatenation, it is a dimensionality
reduction strategy to reduce the computational cost where the output channel number is significantly cut by calculating the average value of each unit, {the $\hat{\text{F}_u^{C2}}$ is a single-channel feature map. After the concatenation of $(AvgPooling(\text{M})$ and $\hat{\text{F}_u^{C2}}$, we use a \textit{CBR} to fuse early semantic features with the GKT-encoded click-point map to generate a position-enhanced attention map.} The output F$_k^{L}$ represents attentions around the target in $U$. We further integrate it into query features by another element-wise product operation:  
\begin{equation}
\begin{aligned}
    \text{F}_u^{LE} &= \text{F}_u^{L} \cdot \text{F}_u^{E} \\
    &= \text{F}_u^{L} \cdot \text{F}^\text{MHCA} \cdot \text{F}_u^{C3}
\end{aligned}
\end{equation}

The enhanced query features F$_u^{LE}$ provide desired weights on both object-specific and contexture regions. The next step is to build the connection between query and reference to pave the way for the downstream detection task. Spatial Attention (SA) from \cite{sun2023cross} is employed to finalize the attentions within the reference domain: 
\begin{equation}
\begin{aligned}
    \text{A}_s &= SpatialAttention(\text{F}_u^{LE}, \hat{\text{F}}_{s}^{C3}),
\end{aligned}
\end{equation}
where $\hat{\text{F}}_{s}^{C3}$ is $\text{F}_{s}^{C3}$ (in Eq. (5)) transferred by normalization. A$_s$ is the final attention result of the matching block. {The Spatial Attention (SA) module is retained to play a complementary role to MHCA. While MHCA focuses on enhancing global semantic alignment between query and reference views, SA is responsible for preserving fine-grained local cues that are essential for precise matching, particularly in cluttered or visually diverse reference scenes. This design choice improves the coherence of the overall matching strategy by balancing cross-view global context with local spatial discriminability.}

\subsection{Detection Head}

Once the final cross-view attention result (i.e. A$_s$) is obtained, the reference features can be weighted accordingly, by using a element-wise product wrapped with CBR for a full fusion. Considering that we are expecting outputs of object bounding boxes, a convolution layer is needed:
\begin{equation}
\begin{aligned}
H = \text{Conv1D} (\textit{CBR}(\hat{\text{F}}_s^{C3} \cdot \text{A}_s)).
\end{aligned}
\end{equation} 
The outputs H include regression results of bounding boxes associated with corresponding classification scores. 9 anchors are used in our work where the anchor with the highest classification score is a prediction. The regression yields the size and center coordinates of each bounding box, and the classification gives the probability that a query object is located at the bounding box.

\begin{table*}[ht]
\centering
\caption{The test result of our method and existing methods on CVOGL}
\label{CVOGL_pre}
\setlength{\tabcolsep}{4pt} 
\begin{tabular}{@{}ccccccccccc@{}}
\hline
\multirow{2}{*}{Data} & \multicolumn{4}{c}{Drone $\rightarrow$ Satellite} & \multicolumn{4}{c}{Ground $\rightarrow$ Satellite} \\

 & \multicolumn{2}{c}{Validation} & \multicolumn{2}{c}{Test} & \multicolumn{2}{c}{Validation} & \multicolumn{2}{c}{Test} \\
\hline
Method & acc@0.25(\%)$\uparrow$ & acc@0.50(\%)$\uparrow$ & acc@0.25(\%)$\uparrow$ & acc@0.50(\%)$\uparrow$ & acc@0.25(\%)$\uparrow$ & acc@0.50(\%)$\uparrow$ & acc@0.25(\%)$\uparrow$ & acc@0.50(\%)$\uparrow$ \\
\hline
CVM-Net & 20.04 & 3.47 & 20.14 & 3.29 &5.09 &0.87 &4.73 &0.51 \\
RK-Net & 19.94 & 3.03 & 19.22 & 2.67 &8.67 &0.98 &7.40 &0.82\\
LR2LTR & 38.68 & 5.96 & 38.95 & 6.27 &12.24 &1.84 &10.69 &2.16 \\
Polar-SAFA & 36.19 & 6.39 & 37.41 & 6.58 &19.18 &2.71 &20.66 &3.19 \\
TransGeo & 34.78 & 5.42 & 35.05 & 6.37 &21.67 &3.25 &21.17 &2.88\\
SAFA & 36.19 & 6.39 & 37.41 & 6.58 &20.59 &3.25 &22.20 &3.08 \\
DetGeo & 59.81 & 55.15 & 61.87 & 57.55 &46.70 &43.99 &45.43 &42.24 \\
VAGeo & 64.25 & 59.59 & 66.19 &  61.87 &47.56 &44.42 &48.21 &45.22 \\
Our & \textbf{66.52} & \textbf{61.86} & \textbf{68.35} & \textbf{63.93} &\textbf{48.54}& \textbf{44.20}&\textbf{51.49}& \textbf{47.69}\\
\hline
\end{tabular}
\end{table*}

In the detection head, the loss function used to evaluate the difference between predictions and ground truth should account for both localization and classification losses, as $\mathcal{L} = \mathcal{L}_{MSE} + \mathcal{L}_{BCE}$ where $\mathcal{L}_{MSE}$ is the mean squared error (MSE) loss, capturing the localization loss, while $\mathcal{L}_{BCE}$ is the binary cross entropy (BCE) loss for classification.

\begin{figure}
    \centering
    {
        \includegraphics[width=1.55in]{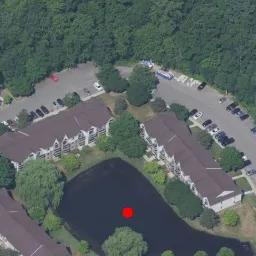}
    }
    {
        \includegraphics[width=1.55in]{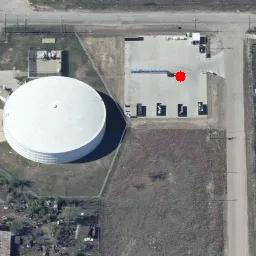}
    } \hfill\\
    {
        \includegraphics[width=1.55in]{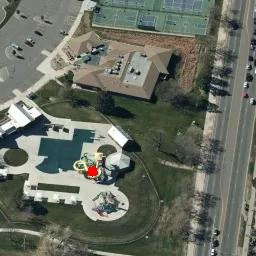}
    }
    {
        \includegraphics[width=1.55in]{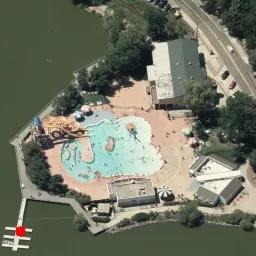}
    }
    \caption{Examples of additional objects in the CVOGL-fewshot dataset: Lake, Parking, Slide, and Port.}
    \label{f4}
\end{figure}

\section{Experimental Results}

\subsection{Dataset Overview and Few-Shot Extension}

The \textbf{CVOGL} dataset is currently the only publicly available dataset for evaluating object-level geo-localization tasks. It consists of 5,279 ground-view images, 5,279 drone-view images, and 5,836 high-resolution satellite images. The dataset primarily includes common objects such as buildings, bridges, roundabouts, baseball fields, and storage tanks. Each object is annotated with a click point in the ground/drone views, a bounding box, and an object label in the satellite view.

To enhance CVOGL and evaluate the few-shot capabilities of our proposed method, we extend the dataset by adding new object categories. For experiments involving few-shot learning, we refer to this extended dataset as \textbf{CVOGL-fewshot}; otherwise, all experiments are conducted on the original CVOGL dataset.

The CVOGL-fewshot dataset contains a total of 28 samples for training and 24 samples for testing, with each of the four new categories represented by 7 training samples. This setup adheres to standard few-shot learning conditions. We construct CVOGL-fewshot mainly for the Drone-to-Satellite task where we labeled additional objects not present in the original CVOGL dataset. This process involved manually annotating four new categories—lake, parking, slide, and port—by aligning OpenStreetMap and satellite images at matching locations and scales to ensure accurate bounding box annotations. Some examples are shown in Fig.~\ref{f4}.

\subsection{Experiment Setting}

\subsubsection{Implementation Details} 

Our framework is implemented in Python, and the experiments were conducted using an NVIDIA A100 GPU with 80GB memory. The same size of images with the same pre-processing are used in all our experiments: $256 \times 256$, $256 \times 512$, and $1024 \times 1024$ for drone, ground, and satellite respectively.

For hyper-parameters, we used a batch size of 12, a learning rate of $1e-4$, and 25 epochs in all CVOGL experiments. In the few-shot experiments on CVOGL-fewshot, the batch size was adjusted to 6, and the number of epochs was set to 20.

In training and evaluating the few-shot task, we all initialized from pre-trained models (checkpoints listed in Table \ref{CVOGL_pre}) and fine-tuned them on the CVOGL-fewshot dataset.

\begin{figure}[h]
\centering
\includegraphics[scale=0.19]{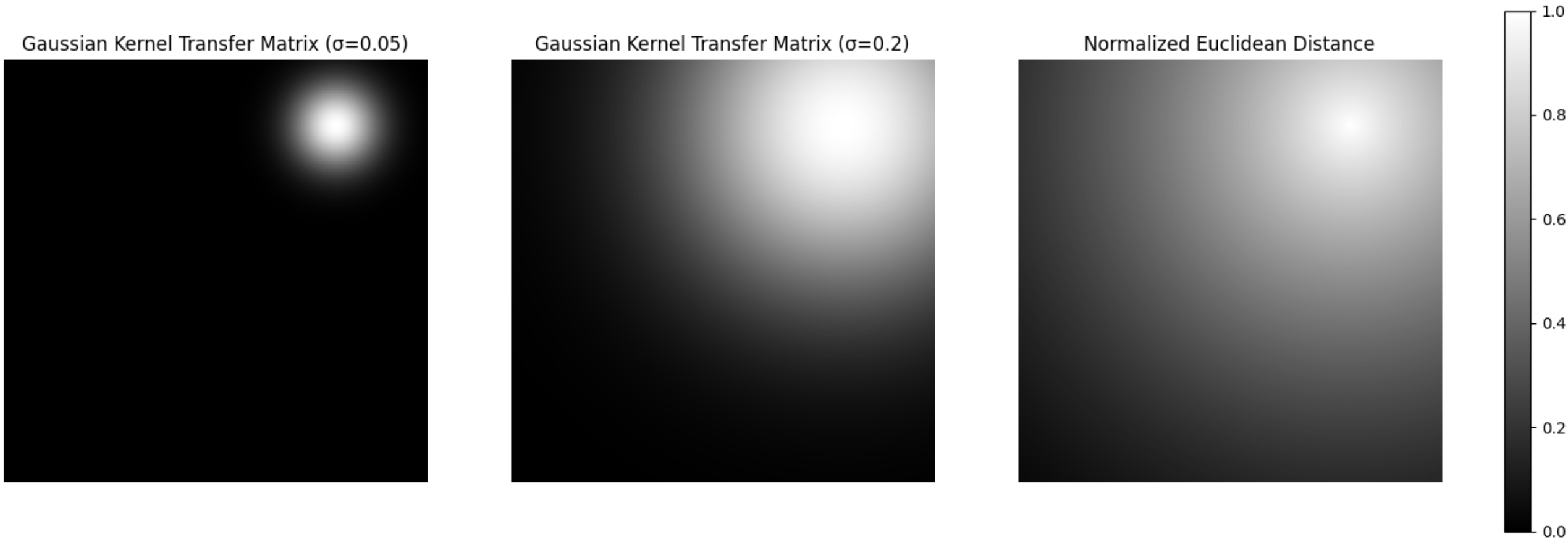}
\caption{Different click-point embedding maps (i.e. \textbf{M} defined in Eq.(1) and Eq.(2)). Left: $\sigma=0.05$, Middle: $\sigma=0.2$, Right: The distance map used in \cite{sun2023cross}.}
\label{cp}
\end{figure}

\begin{figure}[h]
\centering
\includegraphics[scale=0.50]{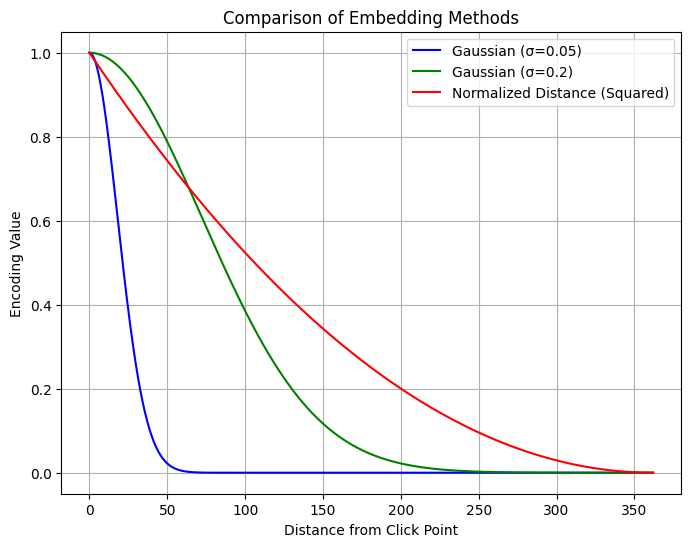}
\caption{Illustration of embedding values in the three maps in Fig.~\ref{cp}.}
\label{f8}
\end{figure}

\subsubsection{Evaluation Setting} 

As this is an object detection task, our evaluation metrics are primarily based on Intersection over Union (IoU) and accuracy. IoU measures the overlap between ground truth (GT) and predicted bounding boxes. The equations are as follows:
\begin{equation}
\begin{aligned}  
    \text{acc@t} = \frac{1}{N} \sum_{i=1}^{N} \psi_i,
\end{aligned}
\end{equation}
\begin{equation}
\begin{aligned}  
      \psi_i(t) = \begin{cases} 1 & \text{if } \text{IoU} \geq t \\ 0 & \text{if } \text{IoU} < t \end{cases},  \\
    \text{IoU}(b_i,b_i^*) = \frac{|b_i \cap b_i^*|}{|b_i \cup b_i^*|},
\end{aligned}
\end{equation}
$b_i$ and $b_i^*$ represent the \textit{i}\textsuperscript{th} instance ground-truth and predicted bounding box. The $\text{IoU}(b_i, b_i^*)$  is the ratio of the overlap area between the predicted bounding box $b_i$ and the ground truth bounding box $b_i^*$ to their union area. $t$ denotes a threshold to distinguish the prediction is correct or not.  {N represents the total number of samples in the test or validation set.} In our experiments, IoU results with the settings of $t = 0.50$ and $t = 0.25$  are reported.  

To provide a clearer indication of detection performance, we also report acc@0.25 and acc@0.50 as additional accuracy metrics. Accuracy is typically defined as the ratio of correctly predicted instances to the total number of instances in a dataset, as below. 
\[
\text{Accuracy} = \frac{\text{Number of Correct Predictions}}{\text{Total Number of Predictions}} \times 100
\]
acc@0.25 is the accuracy (in \%) where the IoU between the predicted bounding box and the ground truth is greater than or equal to 0.25. acc@0.50 is the accuracy (in \%) where the IoU is greater than or equal to 0.50.

\begin{figure}[ht]
\centering
\includegraphics[scale=0.5]{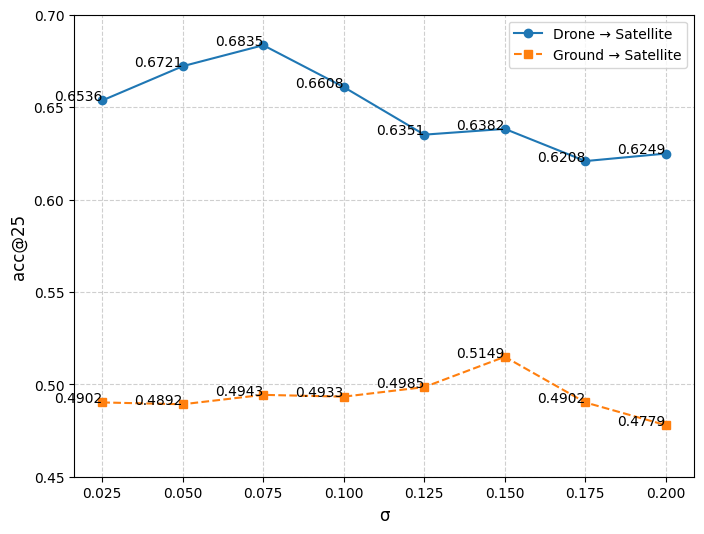}
\caption{Sensitivity analysis of Gaussian standard deviation $\sigma$ in Eq.(1) and Eq.(2).}
\label{sd}
\end{figure}

\begin{figure*}[h]
\centering
\includegraphics[scale=0.43]{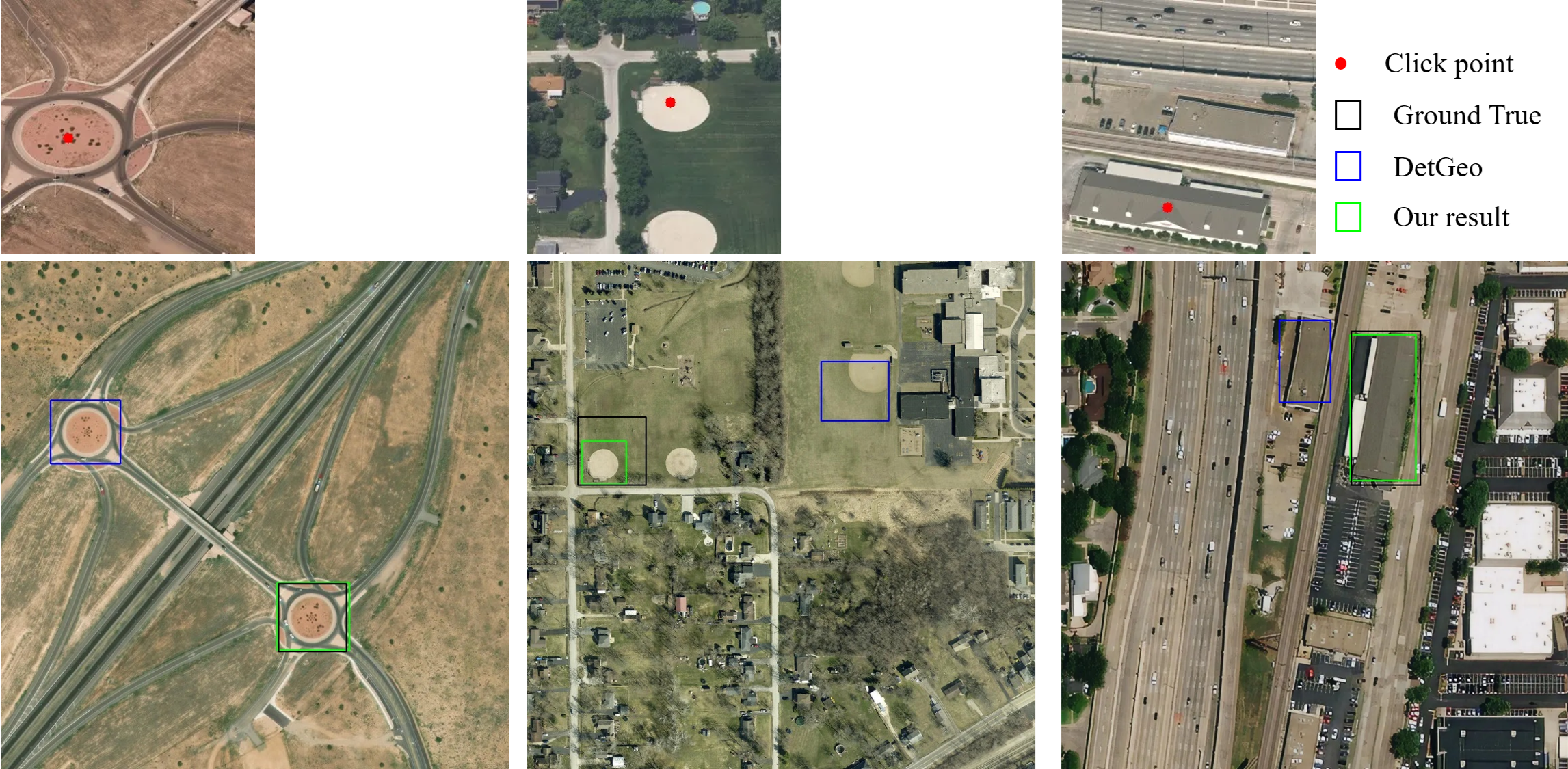}
\caption{Visual comparison of object-level geo-localization results between our model and the previous method. Despite the presence of multiple similar targets in the satellite view, our model (green bounding boxes) accurately localizes the specified target, demonstrating improved precision over the previous method (blue bounding boxes).}
\label{f5}
\end{figure*}

\subsubsection{Parameter Analysis}

The click-point embedding map \textbf{M} determines the extent of the areas that should be involved to accurately capture object-specific information. In accordance with Eq.(1) and Eq.(2), the standard deviation $\sigma$ of GKT is a critical parameter for defining the relevant region of interest. 

Examples of different embedding maps are listed in Fig.~\ref{cp} with embedding values illustrated in Fig.~\ref{f8}.  Comparing GKT with the previous method (i.e. the distance map in \cite{sun2023cross}), we noticed that the GKT curve decreases rapidly, particularly under the condition of $\sigma=0.05$. Benefiting from this characteristic, GKT can focus positional encoding information on the clicked object in query images containing multiple objects, thereby reducing the diffusion of positional information during neural network propagation and enhancing the model's robustness in complex scenarios.

 {To find a proper setting of $\sigma$, experiments with intervals of 0.025, starting at 0.025 and ending at 0.20 have been tested. As shown in Fig.~\ref{sd}, $\sigma = 0.075$ works best in our experiments on the Drone $\rightarrow$ Satellite task, which results in an accuracy of 68.35\% at acc@0.25 on the test set. For the Ground $\rightarrow$ Satellite task, the peak performance occurs at $\sigma = 0.15$, yielding an accuracy of 51.49\%. This suggests that the Drone modality benefits from a more concentrated representation of location information, likely due to the smaller size of the annotated objects in the Drone view. Furthermore, we noticed that variations in $\sigma$ have a significant impact on performance. For the Drone $\rightarrow$ Satellite task, the accuracy difference between $\sigma = 0.075$ and $\sigma = 0.20$ is 5.86\%, while for the Ground $\rightarrow$ Satellite task, the difference between $\sigma = 0.15$ and $\sigma = 0.20$ is 3.70\% at acc@0.25. This observation highlights the importance of selecting the appropriate map $M$ to optimize the model's performance.}

\begin{figure*}[h]
\centering
\includegraphics[scale=0.43]{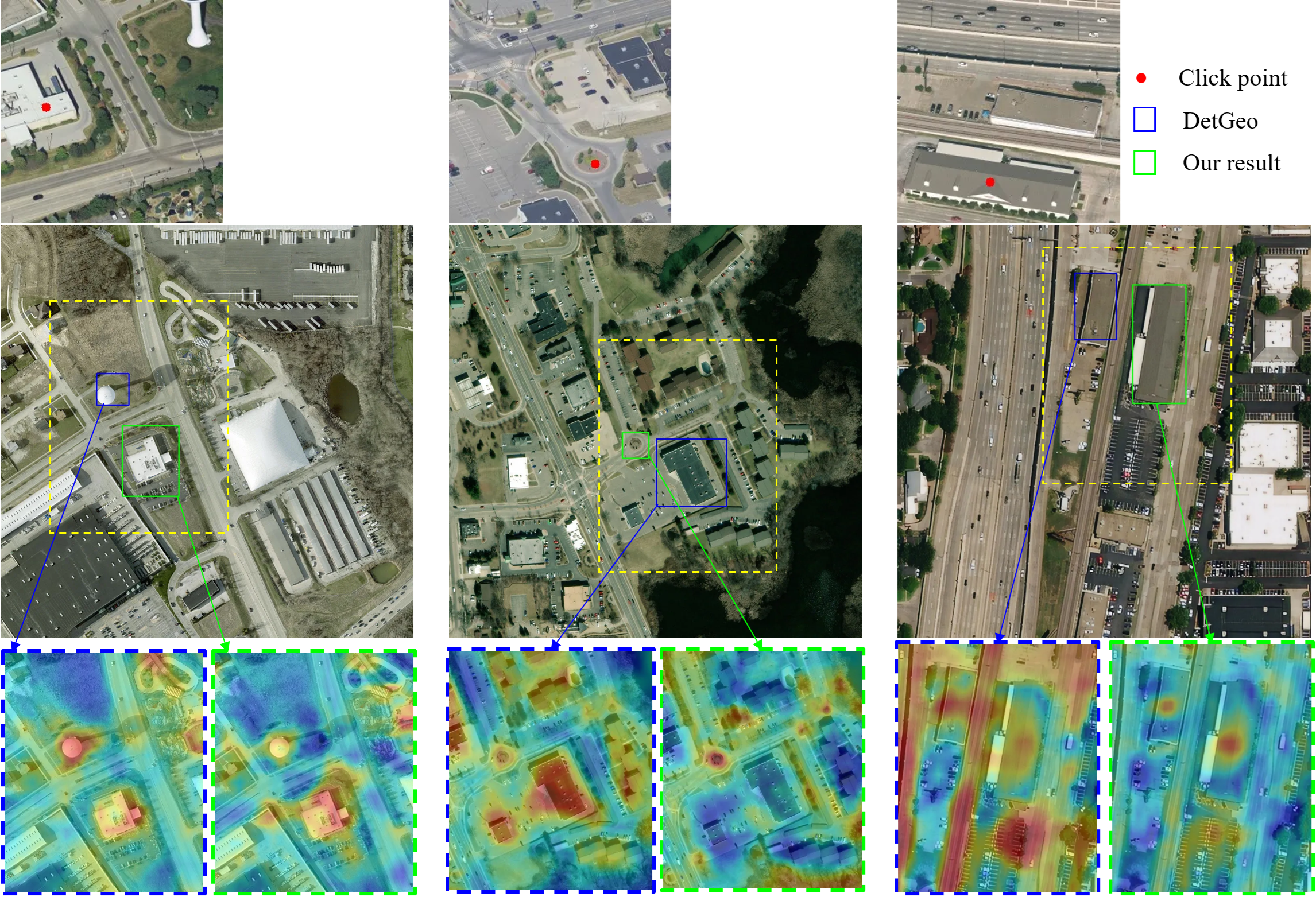}
\caption{ Attention map comparison between DetGeo and our model for object-level geo-localization. The top row shows query images (drone views), the middle row displays reference images (satellite views), and the bottom row provides zoomed-in attention maps within dashed-yellow rectangular areas. Attention maps from DetGeo (dashed-blue rectangular areas) highlight most of common objects shown in the dataset, while our attention maps (dashed-green rectangular area) focuse more selectively on the target object specified by the user, enhancing localization accuracy.}
\label{f6}
\end{figure*}

\subsection{Comparison}

Our developed OCGNet has been compared with the following methods on the CVOGL dataset: CVM-Net \cite{hu2018cvm}, SAFA \cite{shi2019spatial}, RK-Net \cite{lin2022joint}, L2LTR \cite{yang2021cross}, TransGeo \cite{zhu2022transgeo}, DetGeo \cite{sun2023cross}, and VAGeo\cite{li2025vageo}. In addition, we compare OCGNet with DetGeo on the CVOGL Few-shot dataset. 

Since CVOGL presents a novel challenge in remote sensing, only DetGeo focuses on cross-view geo-localization via object detection. Therefore, the primary comparison is between OCGNet and DetGeo on both CVOGL and CVOGL-fewshot. The results of other methods such as CVM-Net, SAFA, RK-Net, L2LTR, and TransGeo are directly from \cite{sun2023cross}.

\subsubsection{Overview of Performance Comparison} 

{In Table \ref{CVOGL_pre}, we report a comprehensive comparison between our method and a series of models on the CVOGL task, which consists of two query types: Drone $\rightarrow$ Satellite and Ground $\rightarrow$ Satellite. As shown in the table, our method consistently surpasses all existing methods across all evaluation metrics. For the Drone $\rightarrow$ Satellite task, our method achieves the highest performance with 68.35\% acc@0.25 and 63.93\% acc@0.50, outperforming the previous best end-to-end method (VAGeo) by 2.16\% and 2.06\% respectively. On the more challenging Ground $\rightarrow$ Satellite task, our approach reaches 51.49\% acc@0.25 and 47.69\% acc@0.50, marking a maximum improvement of 3.28\% over VAGeo. These results clearly demonstrate the superiority and robustness of our method under both cross-view scenarios. Moreover, OCGNet improved on the DetGeo baseline by \textbf{6.48\%} and \textbf{6.06\%} for the ground $\rightarrow$ satellite task.}

\begin{table}[h]
    \centering
    \caption{Few-shot learning performance comparison}
    \label{CVOGL_few}
    \begin{tabular}{cccc}
        \hline
        Method & acc@0.25(\%)$\uparrow$ & acc@0.50(\%)$\uparrow$ & IoU(\%)$\uparrow$ \\
        \hline
        DetGeo & 16.67 &16.67&13.70\\
        Our & 29.17 & 25.0 & 20.18 \\
        \hline
    \end{tabular}
\end{table}

In Table \ref{CVOGL_few}, we present a comparison of the few-shot performance between our model and the state-of-the-art method (DetGeo). On the CVOGL Drone $\rightarrow$ Satellite few-shot task, our model achieves 29.17\% acc@0.25 and 25.0\% acc@0.50, improving by 12.5\% in acc@0.25 and 8.33\% in acc@0.50. These results demonstrate the generalizability our model from limited training samples (i.e. 7), highlighting its strong potentials in few-shot scenarios.

\begin{figure}[h]
\centering
\includegraphics[scale=0.34]{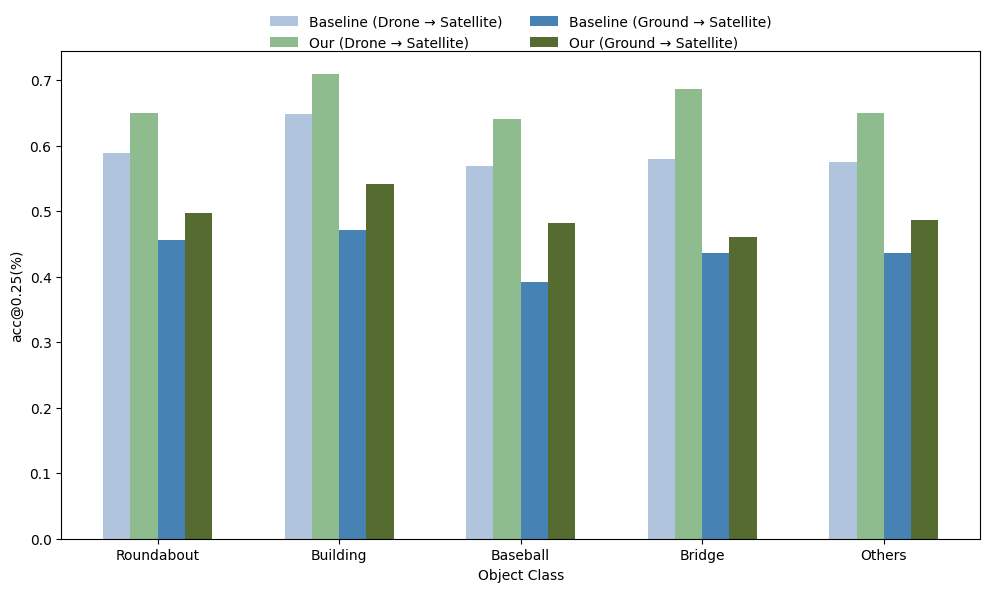}
\caption{Performance comparison across different object classes for both Drone $\rightarrow$ Satellite and Ground $\rightarrow$ Satellite tasks.}
\label{class}
\end{figure}

\subsubsection{Visual Comparison}

Fig.~\ref{f5} presents a set of results comparing our model with the previous method, DetGeo. These examples are particularly challenging, as the satellite views contain multiple similar targets, making object-level geo-localization more complex. As shown in the results, our model effectively identifies the specific target amidst similar objects, demonstrating superior localization accuracy. This improvement can be attributed to enhanced query features achieved through 1) the LE module, which preserves object-specific information during the matching stage, and 2) the MHCA module, which selectively focuses on relevant regions surrounding the object.

\begin{table}[ht]
    \centering
    \caption{Number of parameters comparison}
    \label{parameters}
    \begin{tabular}{ccc}
        \hline
        Method & Number of parameters & Average inference time\\
        \hline
        DetGeo & 73.8M & 15 ms\\
        Our & 74.8M & 16 ms\\
        \hline
    \end{tabular}
\end{table}

To further illustrate these differences, we visualize intermediate results using attention maps, as seen in Fig.~\ref{f6}. The detection results from DetGeo and our model are marked with blue and green bounding boxes, respectively. The top rows display the query images (i.e., drone views), while the middle rows show the reference images (i.e., satellite views). The bottom row zooms into the attention maps of a local area (highlighted by a dashed-yellow box) for both DetGeo and our model, marked with dashed blue and green colors, respectively. Comparing each pair of attention maps, it becomes evident that DetGeo tends to highlight all common objects (such as towers, buildings, roundabouts, etc.), whereas our model provides more targeted attention to the specific objects indicated by the user's click point. {Additionally, the attention maps reveal that OCGNet performs better in complex scenarios involving multiple objects.}

\subsubsection{Performance in Different Objects}

We evaluated the performance of our model across five object classes on CVOGL, as presented in Fig.~\ref{class}. As mentioned before, DetGeo used for comparison represents the current state-of-the-art method in the object-level geo-localization task. 

From the results shown in Fig.~\ref{class}, it can been seen that our model consistently outperforms DetGeo across the five object classes, demonstrating a notable improvement in accuracy and robustness. This improvements are attributed to the effective leverage of both object-specific and contexture information. Our model preserves critical information during the matching process, allowing for accurate distinction among visually similar objects. Additionally, its adaptability across different viewpoints (Drone $\rightarrow$ Satellite and Ground $\rightarrow$ Satellite) further highlights its resilience to changes in scale, angle, and visual complexity. These results underscore the versatility and generalization capability of our model, providing a reliable framework for diverse object-level geo-localization scenarios.

{Noting that the \textit{Baseball} and \textit{Bridge} classes have relatively small training datasets (292 and 238 samples, respectively) compared to the \textit{Building} class with 2175 samples, Fig.~\ref{class} shows that our method achieves significant improvements of 7.17\% and 10.67\% respectively, in the Drone $\rightarrow$ Satellite task. In the Ground $\rightarrow$ Satellite task, the \textit{Baseball} class outperforms the baseline with a 9.03\% improvement. These findings further demonstrated the robustness of our approach to scenarios with limited training data. }

\subsubsection{Model Parameters Comparison}

The number of learnable parameters is an important factor when assessing the computational cost of a model. As shown in Table \ref{parameters}, our model and DetGeo have comparable numbers of learnable parameters, with 74,845,918 and 73,795,164 parameters, respectively. All experiments, along with the Gradio demo available on our GitHub, confirm that the training and inference costs are nearly identical when using either the GTX 4090 or A100 GPU. For inference, we ran our model and DetGeo 100 times each, with average inference times of 16 milliseconds (ms) for our model and 15 milliseconds (ms) for DetGeo.

\subsection{Ablation Study}

To evaluate the contribution of each component in our proposed framework, we perform a series of ablation experiments. As summarized in Table~\ref{ablation}, we examine the following configurations: 1) the baseline DetGeo, 2) - 4) adding each of the LE, MHCA, or GKT modules individually, and 5) - 7) removing each module one at a time from the full model. These configurations help isolate the effects of LE, MHCA, and GKT modules, which are integrated into our framework based on the DetGeo baseline.

\begin{table*}[ht]
\centering
\caption{Ablation study of different blocks on CVOGL}
\label{ablation}
\setlength{\tabcolsep}{3pt} 
\begin{tabular}{@{}lccc|cccc@{}} 
\hline
\multirow{2}{*}{Method} & \multirow{2}{*}{LE} & \multirow{2}{*}{MHCA} & \multirow{2}{*}{GKT} & \multicolumn{2}{c}{Drone $\rightarrow$ Satellite} & \multicolumn{2}{c}{Ground $\rightarrow$ Satellite} \\

 & & & & acc@0.25(\%)$\uparrow$ & acc@0.50(\%)$\uparrow$ & acc@0.25(\%)$\uparrow$ & acc@0.50(\%)$\uparrow$ \\
\hline
1) DetGeo & \ding{55} & \ding{55} & \ding{55} & 61.87 & 57.55 & 45.43 & 42.24 \\
2) LE & \checkmark & \ding{55} & \ding{55} & 63.82 & 57.25 & 46.56 & 42.96 \\
3) MHCA & \ding{55} & \checkmark & \ding{55} & 62.05 & 56.60 & 47.48 & 44.19 \\
4) GKT & \ding{55} & \ding{55} & \checkmark & 66.29 & 60.95 & 47.07 & 43.68 \\
5) No LE & \ding{55} & \checkmark & \checkmark & 63.10 & 57.97 & 45.22 & 41.73 \\
6) No MHCA & \checkmark & \ding{55} & \checkmark & 66.19 & 60.32 & 49.54 & 45.22 \\
7) No GKT & \checkmark & \checkmark & \ding{55} & 64.03 & 59.61 & 48.72 & 45.73\\
Our model & \checkmark & \checkmark & \checkmark & \textbf{68.35} & \textbf{63.93} & \textbf{51.49} & \textbf{47.69}\\
\hline
\end{tabular}
\end{table*}

Based on the experiment in Table \ref{ablation}, we obtain several observations as follows:
\begin{itemize}
    \item Individual Module Effectiveness (i.e. the experiments 2), 3) and 4)): Each module (i.e. LE, MHCA, and GKT) brings improved accuracy over the DetGeo baseline. GKT shows the most significant gains with improving Drone $\rightarrow$ Satellite accuracy by 4.42\% and Ground $\rightarrow$ Satellite accuracy by 1.64\%, which confirm its strong impact on precise localization. These improvements validate the utility and generalizability of all three modules.
    
    \item Module Contributions (i.e. the experiments 5), 6) and 7)): When removing modules from the full model, performance drops noticeably. Excluding MHCA at 6) affects the Ground $\rightarrow$ Satellite task the most, while excluding GKT at 7) or LE at 5) more significantly impacts Drone $\rightarrow$ Satellite accuracy. This suggests MHCA is more critical for complex ground-level viewpoints, whereas GKT and LE enhance localization in drone imagery.
    
    \item GKT vs. Euclidean Map (i.e. the experiment 4) vs. 7)): Using GKT to generate the click-point embedding map \textbf{M} yields better results than using Euclidean distance maps, particularly for Drone $\rightarrow$ Satellite. This may be due to smaller object scales in drone views, where GKT enables more focused attention on the relevant object.
    
\end{itemize}

In summary, the ablation results highlight the complementary roles of LE, MHCA, and GKT in improving performance. Their integration leads to significant gains in accuracy across both Ground $\rightarrow$ Satellite and Drone $\rightarrow$ Satellite tasks, confirming their effectiveness and synergy in object-level cross-view geo-localization.

\section{Conclusion} 
We present OCGNet, a novel Object-level Cross-view Geo-localization Network designed for precise localization of visually similar objects in UAV and ground imagery using satellite references. By integrating location information twice and enhancing query features through GKT, LE and MHCA, OCGNet significantly boosts localization accuracy, achieving state-of-the-art performance on the CVOGL dataset.

OCGNet also demonstrates strong few-shot generalization, making it practical for real-world scenarios with limited annotated data, such as search-and-rescue and infrastructure monitoring. The key contributions of dual-location integration, query feature enhancement and generalization highlight its potential for advancing object-level geo-localization. Our current evaluation is constrained by relying on the CVOGL dataset, which uniquely offers cross-view imagery with click-point annotations. As future work, we aim to develop more comprehensive datasets tailored for few-shot object-level geo-localization.

\bibliographystyle{IEEEtran}
\bibliography{refs.bib}

\end{document}